\documentclass{article}
\usepackage[latin1]{inputenc}
\usepackage{amssymb, amsmath, amsthm}
\usepackage[a4paper]{geometry}
\usepackage{authblk} 
\usepackage{pifont}
\usepackage{graphicx}
\usepackage{xtab} 
\usepackage{longtable} 
\usepackage{footnote}
\makesavenoteenv{tabular}
\usepackage{tabularx}
\usepackage{rotating}
\usepackage{booktabs}
\usepackage[dvipsnames,table]{xcolor}
\newcolumntype{R}{>{\raggedleft\arraybackslash}X}
\newcolumntype{L}{>{\raggedright\arraybackslash}X}
\newcolumntype{C}{>{\centering\arraybackslash}X}

\usepackage{dcolumn} 
\newcolumntype{.}{D{.}{.}{-1}}
\usepackage{pdflscape}
\usepackage{subcaption}
\usepackage{afterpage}
\usepackage{capt-of}%

\usepackage{pgfplots}
\usepackage{tikz}
\usetikzlibrary{arrows,decorations.markings}
\usepackage{url}
\usepackage{natbib}
\usepackage{hyperref}
\hypersetup{
  colorlinks   = true,    
  urlcolor     = blue,    
  linkcolor    = blue,    
  citecolor    = red      
}

\usepackage{enumitem}

\setlist[itemize]{leftmargin=3\parindent}
\setlist[enumerate]{leftmargin=2\parindent}

\theoremstyle{plain}

\newtheorem{definition}{Definition}[section]

\newtheorem{theorem}{Theorem}[section]

\def\keywords{\vspace{.5em} 
{\textit{Keywords}:\,\relax%
}}

\author{S\'andor Boz\'oki\thanks{Institute for Computer Science and Control, Hungarian Academy of Sciences (MTA SZTAKI),
Laboratory on Engineering and Management Intelligence, Research Group of Operations Research and Decision Systems and Department of Operations Research and Actuarial Sciences, Corvinus University of Budapest, Hungary \\ e-mail: bozoki.sandor@sztaki.mta.hu} --
L\'aszl\'o Csat\'o\thanks{Department of Operations Research and Actuarial Sciences, Corvinus University of Budapest and MTA-BCE ''Lend\"ulet'' Strategic Interactions Research Group, Hungary \\ e-mail: laszlo.csato@uni-corvinus.hu} --
J\'ozsef Temesi\thanks{Department of Operations Research and Actuarial Sciences, Corvinus University of Budapest, Hungary \\ e-mail: jozsef.temesi@uni-corvinus.hu} 
}
\title{An application of incomplete pairwise comparison matrices for ranking top tennis players}
\date{\today}

\begin{document}

\maketitle

\begin{abstract}
Pairwise comparison is an important tool in multi-attribute decision making. Pairwise comparison matrices (PCM) have been applied for ranking criteria and for scoring alternatives according to a given criterion. Our paper presents a special application of incomplete PCMs: ranking of professional tennis players based on their results against each other. The selected 25 players have been on the top of the ATP rankings for a shorter or longer period in the last 40 years. Some of them have never met on the court. One of the aims of the paper is to provide ranking of the selected players, however, the analysis of incomplete pairwise comparison matrices is also in the focus. The eigenvector method and the logarithmic least squares method were used to calculate weights from incomplete PCMs. In our results the top three players of four decades were Nadal, Federer and Sampras. Some questions have been raised on the properties of incomplete PCMs and remains open for further investigation.

\end{abstract}

\section{Introduction}

A well-known application field of pairwise comparison matrices (PCMs) is multi-attribute decision making (MADM). The values of pairwise comparisons are applied for ranking of criteria or for scoring alternatives to a given criterion.

This paper will use pairwise comparison values for ranking of tennis players based on their results against each other. Our aim is to make a 'historical' comparison of top tennis players of the last 40 years. The ranking idea is how the players performed against each other in a pairwise manner in the long run. We have collected the results of 25 players who have been on the top of the ATP ranking lists for a shorter or a longer period.\footnote{~One can ask, why not an all-time ranking? The answer is simple: our data collection used the official ATP website. The ATP database contained reliable and complete data from 1973 (see at \url{http://www.atpworldtour.com/Players/Head-To-Head.aspx}).}

There could be several reasons why some elements of a $PCM$ are missing. It can happen that decision makers do not have time to make all comparisons, or they are not able to make some of the comparisons. Some data could have lost, but it is also possible that the comparison was not possible. In our case the reason of missing elements is obvious: we are not able to compare those players directly who have never played against each other.

Professional tennis is very popular around the world. The professional tennis associations (ATP, WTA) have been collecting data about the tournaments and the players. There is a freely available database about the results of the top tennis players including data from 1973. That gave the possibility to construct the pairwise comparison matrices of those players who have been leading the ATP ranking for a period of any length. Applying one of the estimation methods for generating a weight vector we can produce an order of the players: a ranking. That approach might be highly disputable among tennis fans, of course, but we have to note that other ranking ideas are also based on consensus or tradition, and there is no unique answer to the question 'Who is the best?'.

The existing ATP rankings, for instance, give points to the players for certain periods according to the importance of the ATP tournaments (based on the prize money) using simple rules for correcting the impacts of some biasing conditions.

The media and most of the experts consider \#1 of the ATP-ranking as the 'best' tennis player. Our approach is also ranking-oriented, but we will not use this term, the emphasis will be put on the excellence of players with higher positions relative to those who have lower ranking positions.
Ranking of players will be done according to the weights, and the player with the highest weight can be regarded as the 'best', however, this term is restricted to our sample of players and varies as different ranking lists are generated.

In recent years some papers have attempted to rank professional tennis players with the use of well-founded methods. \citet{Radicchi2011} considered all matches played between 1968 and 2010 to construct a weighted and directed preference graph. It develops a diffusion algorithm similar to Google's PageRank \citep{BrinPage1998} to derive the ranking of nodes representing the tennis players. It also provides lists for specific playing surfaces and different time periods. On the basis of the whole dataset, \emph{Jimmy Connors} was identified as the \#1 player. He is also the winner of the decade 1971-80. For subsequent years, the \#1 players are \emph{Ivan Lendl} (1981-1990), \emph{Pete Sampras} (1991-2000) and \emph{Roger Federer} (2001-2010). The new ranking has a higher predictive power than the official ATP ranking and does  not require arbitrary external criteria, with the exception of a control parameter.

\citet{DingleKnottenbeltSpanias2013} use this method to derive PageRank-based tennis rankings instead of the official ATP and WTA rankings. For top-ranked players, they are broadly similar, but there is a wide variation in the tail. The PageRank-based rankings are found to be better predictor of match outcomes. \citet{SpaniasKnottenbelt2013} present two new algorithms, SortRank and LadderRank, which make use of a quantitative tennis model to assess the performance of players and compare them with each other.
\citet{Dahl2012} introduce a parametric method based on linear algebra considering the importance of the matches, too.
\citet{MotegiMasuda2012} propose a network-based dynamical ranking system, taking into account that the strength of players depend on time. The method outperforms both the official ranking and \citet{Radicchi2011}'s prestige score in prediction accuracy.

Several authors build statistical models with the aim of a good prediction power. \citet{ClarkeDyte2000} argue that since the rankings are derived from a points rating, an estimate of each player's chance in a head to head contest can be made from the difference in the players' rating points. Using a year's tournament results, a logistic regression model can be fitted to the ATP ratings to estimate that chance.
\citet{McHaleMorton2011} apply a Bradley-Terry type model \citep{BradleyTerry1952} to obtain forecasts, and they show that these forecasts are more accurate according to several criteria than the forecasts obtained from standard models employed in the literature. They compare the model to two logit models, one using official rankings and another using the official ranking points of the two competing players.
\citet{IronsBuckleyPaulden2014} refine that model to be more transparent, fair and insensitive to bias. As they say, even the simplest model improves significantly over the current system, despite having three of the same constraints: no surface information is used, only match results count, and a 12 month rolling window is used to weight games.

\citet{RuizPastorPastor2013} apply Data Envelopment Analysis. According to their model, the 'efficient' players can be used for the 'inefficient' ones as benchmark in order to improve certain characteristics of their play. The ranking is based on cross-efficiency ratios.



Our paper discusses some theoretical results and applications of the incomplete pairwise comparison matrices. This section describes the aim of our research and reviews sport applications with a focus on tennis rankings. The ranking approach implies the use of pairwise comparisons in a natural way. Section~\ref{Sec2} provides an overview of the results in the area of incomplete pairwise comparison matrices -- some of them have been published previously by the authors of this paper. The applied model for top professional tennis players is introduced in the first part of Section \ref{Sec3} together with the description of the database and methodology. The second part of Section~\ref{Sec3} describes the derived rankings -- the Eigenvector Method and the Logarithmic Least Squares Method are applied --, and analyzes some properties of these results. Section~\ref{Sec4} includes further analysis and draws conclusions with some remaining open questions.

\section{Theory and methods} \label{Sec2}

Our paper applies the method of pairwise comparisons.

\begin{definition} \label{Def21}
\emph{Pairwise comparison matrix}:
Let ${\mathbb{R}}_{+}^{n \times n}$ denote the class of $n \times n$ matrices with positive real elements.
The matrix
\begin{center}
$ \mathbf{A}=
\begin{pmatrix}
     1     &    a_{12}  &  a_{13}  & \ldots & a_{1n}   \\
1/{a_{12}} &       1    &  a_{23}  & \ldots & a_{2n}   \\
1/{a_{13}} & 1/{a_{23}} &     1    & \ldots & a_{3n}   \\
    \vdots &    \vdots  &  \vdots  & \ddots & \vdots   \\
1/{a_{1n}} & 1/{a_{2n}} &1/{a_{3n}}& \ldots &   1    \\
\end{pmatrix}  \in {\mathbb{R}}_{+}^{n \times n} $
\end{center}
is called a \emph{pairwise comparison matrix}, if 
\[
a_{ii} = 1 \qquad \text{and}    \qquad a_{ij} = \frac{1}{a_{ji}}
\]
for all indices $i,j=1,\dots,n$.
\end{definition}

In our case the alternatives are tennis players. Choosing any two of them ($P_i$ and $P_j$), we have the results of all matches have been played between them. Let the number of winning matches of $P_i$ over $P_j$ be $x$, and the number of lost matches $y$. We can construct the ratio $x_i/y_i$: if it is greater than $1$, we can say that $P_i$ is a 'better' player than $P_j$. In case of $x_i/y_i$ is equal to $1$ we are not able to decide who is the better. Let the $a_{ij}$ element of the matrix $A$ be $x_i/y_i$, and the $a_{ji}$ element be $y_i/x_i$ for all $i,j=1,\dots,n, \, i \neq j$. Choose the diagonal elements $a_{ii} = 1$ for all $i = 1,2, \dots ,n$, thus $\mathbf{A}$ becomes a pairwise comparison matrix according to Definition~\ref{Def21}.

The PCM matrix $\mathbf{A}$ is used to determine a weight vector $\mathbf{w} = (w_1, w_2, \dots , w_n), \, w_i > 0, \, (i = 1, \dots , n)$, where the elements $a_{ij}$ are estimated by $w_i/w_j$. Since the estimated values are ratios, it is a usual normalization condition that the sum of the weights is equal to $1$: $\sum_{i=1}^n w_i = 1$. That estimation problem can be formulated in several ways.
Saaty \citep{Saaty1980} formulated an eigenvalue problem in the Analytic Hierarchy Process ($AHP$), where the components of the right eigenvector belonging to the maximal eigenvalue ($\lambda_{\max}$) of matrix $\mathbf{A}$ will give the weights. We will refer to that procedure as the Eigenvector Method ($EM$).

For solving the estimation problem it could be obvious to apply methods based on distance minimization, too. That approach will estimate the elements of the $\mathbf{A}$ matrix with the elements of a matrix $\mathbf{W}$, where the element $w_{ij}$ of $\mathbf{W}$ is $w_i / w_j, \, w_i$ and $w_j > 0, \, (i,j = 1, \dots , n)$, and the objective function to be minimized is the distance of the two matrices.
\citet{ChooWedley2004} categorized the estimation methods and found $12$ different distance minimization methods of deriving $\mathbf{w}$ from $\mathbf{A}$ based on minimizing the absolute deviation $|a_{ij} - w_i / w_j|$ or $|w_j a_{ij} - w_i|$, or minimizing the square $(a_{ij} - w_i / w_j)^2$ or $(w_j a_{ij} - w_i)^2$.
The effectiveness of some methods has been studied by \citet{Lin2007}.
We will use the Logarithmic Least Squares Method ($LLSM$) \citep{CrawfordWilliams1985, DeGraan1980, Rabinowitz1976}.


Several authors deal with the problem of inconsistency in AHP (see e.g. \citet{BanaeCostaVansnick2008}). In our tennis application intransitivity may occur, therefore inconsistency is a natural phenomenon. However, the data set is given, consistency correction of the matrix elements could not be done.

PCMs may be \emph{incomplete}, that is, they have missing entries
denoted by $*$:
\begin{equation} \mathbf{A}=
\begin{pmatrix}
     1     &    a_{12}  &   \ast   & \ldots & a_{1n}   \\
1/{a_{12}} &       1    &  a_{23}  & \ldots & \ast \\
    \ast   & 1/{a_{23}} &     1    & \ldots & a_{3n}   \\
    \vdots &    \vdots  &  \vdots  & \ddots & \vdots   \\
1/{a_{1n}} &    \ast    &1/{a_{3n}}& \ldots &   1    \\
\end{pmatrix}.
\end{equation}

Main results have been discussed by Harker \citep{Harker1987}, Carmone, Kara and Zanakis \citep{CarmoneKaraZanakis1997},
Kwiesielewicz and van Uden \citep{Kwiesielewicz1996,KwiesielewiczVanUden2003}, Shiraishi, Obata and Daigo \citep{ShiraishiObataDaigo1998,ShiraishiObata2002}, Takeda and Yu \citep{TakedaYu1995}, Fedrizzi and Giove \citep{FedrizziGiove2007}.

\begin{definition} \label{Def22}
\emph{Graph representation of a PCM}: Undirected graph $G := (V,E)$ represents the incomplete pairwise comparison matrix $\mathbf{A}$ of size $n \times n$ such that $V = \{1, 2, \ldots, n \}$ the vertices correspond to the objects to compare and $E = \{ e(i,j) \, | \, a_{ij} \text{ is given and } i \neq j  \}$, that is, the edges correspond to the known matrix elements.
\end{definition}

There are no edges corresponding to the missing elements in the matrix.

Kwiesielewicz \citep{Kwiesielewicz1996} have considered the Logarithmic Least Squares Method ($LLSM$) for incomplete matrices as
\begin{align}
\min \sum \limits_{
             \begin{array}{c}
              (i,j): a_{ij} \text{ is given}    \\
             \end{array}} &
\left[\log a_{ij} -\log\left(\frac{w_{i}}{w_{j}}\right)\right]^2
  \label{eq:IncompleteLLSMProblem-ObjFunction}  \\
\sum\limits_{i=1}^{n}w_{i} &= 1,    \label{eq:IncompleteLLSMProblem-Normalization} \\
w_{i} &> 0, \qquad i=1,2,\dotsc,n.
\label{eq:IncompleteLLSMProblem-Positivity}
\end{align}

\begin{theorem} \citep[Theorem 4]{BozokiFulopRonyai2010}
Optimization problem \eqref{eq:IncompleteLLSMProblem-ObjFunction}-\eqref{eq:IncompleteLLSMProblem-Positivity}
has a unique solution if and only if \emph{G} is connected. Furthermore, the optimal solution is calculated by solving a system of linear equations.
\end{theorem} 

Note that the incomplete $LLSM$ problem asks for the weights, however, missing elements can be calculated as the ratio of the corresponding optimal weights. We will focus only on the weights.

The generalization of the eigenvector method to the incomplete case requires two steps. First, positive variables $x_1, x_2, \ldots, x_d$ are written instead of missing elements as follows:
\begin{equation} \mathbf{A}(\mathbf{x})= \mathbf{A}(x_1, x_2, \ldots, x_d)=
\begin{pmatrix}
     1     &    a_{12}  &    x_1   & \ldots & a_{1n}   \\
1/{a_{12}} &       1    &  a_{23}  & \ldots &   x_d   \\
1/{x_{1}}  & 1/{a_{23}} &     1    & \ldots & a_{3n}   \\
    \vdots &    \vdots  &  \vdots  & \ddots & \vdots   \\
1/{a_{1n}} & 1/{x_{d}}  &1/{a_{3n}}& \ldots &   1    \\
\end{pmatrix}, \label{eq:IncompleteMatrix-x1-xd}   
\end{equation}
Let $\mathbf{x} = (x_1, x_2, \ldots, x_d)^T \in \mathbb{R}^{d}_{+}$.

Saaty \citep{Saaty1980} defined inconsistency index $CR$ as a positive linear transformation of $\lambda_{\max}(\mathbf{A})$ such that $CR(\mathbf{A}) \geq 0$ and $CR(\mathbf{A}) = 0$ if and only if $\mathbf{A}$ is consistent.
The idea that larger $\lambda_{\max}$ indicates higher $CR$ inconsistency led Shiraishi, Obata and Daigo \citep{ShiraishiObataDaigo1998,ShiraishiObata2002} to consider the eigenvalue optimization problem
\begin{equation}
\underset{{\mathbf{x > 0}}}{\min} \, \lambda_{max}(\mathbf{A}(\mathbf{x})).
\label{eq:lambda_max-OptimizationProblem}   
\end{equation}
in order to find a completion that minimizes the maximal eigenvalue, or, equivalently, $CR$. As in case of incomplete $LLSM$, uniqueness is closely related to the connectedness of $G$.

\begin{theorem} \citep[Theorem 2, Corollary 2 and Section 5]{BozokiFulopRonyai2010}
Optimization problem \eqref{eq:lambda_max-OptimizationProblem} has a unique solution if and only if $G$ is connected. Furthermore, \eqref{eq:lambda_max-OptimizationProblem} can be transformed to a convex optimization problem that can be solved efficiently.
\end{theorem}

Second step is to apply the eigenvector method to the completed pairwise comparison matrix.

Parallel with publishing the first theoretical results on incomplete $PCM$s our research team have been seeking for applications. The world of sports provided us a prosperous experimental field. \citet{Csato2013a} has analysed the chess olympiad. A research paper was published later as a chapter in a book in Hungarian \citep{TemesiCsatoBozoki2012} on ranking tennis players. Some early results have been published and some research questions have been formulated there. The recent article expands the scope of the research and reports new results.

\section{Calculation and results} \label{Sec3}

Our aim is to demonstrate that it is possible to compare players from a long period of time. There are several options how to choose from the list of professional players included in the ATP database. All choices have pros and cons. There is no ideal set of players and comparative time periods, because experts have disputes on controversial issues: Who can represent a certain era? May we compare results from different periods of the carrier path of an individual player? May we set up a unified ranking or it is better to have separate rankings for different surfaces?

\begin{figure}[h]
\centering
\caption{Length of professional tennis career for the chosen players}
\label{Fig1}

\begin{tikzpicture}
\begin{axis}[width=\textwidth, 
height=0.8\textwidth,
xmajorgrids=true,
symbolic y coords={Djokovic,Nadal,Roddick,Federer,Ferrero,Hewitt,Safin,Kuerten,Moya,Rios,Kafelnikov,Rafter,Courier,Sampras,Agassi,Muster,Becker,Edberg,Wilander,Lendl,McEnroe,Borg,Connors,Nastase,Newcombe},
ytick=data,
y tick label style = {font=\tiny,color=white},
axis y line*=right,
extra y ticks={Djokovic,Nadal,Roddick,Federer,Ferrero,Hewitt,Safin,Kuerten,Moya,Rios,Kafelnikov,Rafter,Courier,Sampras,Agassi,Muster,Becker,Edberg,Wilander,Lendl,McEnroe,Borg,Connors,Nastase,Newcombe},
extra y tick labels={Djokovic,Nadal,Roddick,Federer,Ferrero,Hewitt,Safin,Kuerten,Moya,Rios,Kafelnikov,Rafter,Courier,Sampras,Agassi,Muster,Becker,Edberg,Wilander,Lendl,McEnroe,Borg,Connors,Nastase,Newcombe},
every extra y tick/.style={
tick0/.initial=blue,tick2/.initial=blue,tick4/.initial=blue,tick6/.initial=blue,tick8/.initial=blue,tick10/.initial=blue,
tick12/.initial=blue,tick14/.initial=blue,tick16/.initial=blue,tick18/.initial=blue,tick20/.initial=blue,tick22/.initial=blue,tick24/.initial=blue,
tick1/.initial=brown,tick3/.initial=brown,tick5/.initial=brown,tick7/.initial=brown,tick9/.initial=brown,tick11/.initial=brown,
tick13/.initial=brown,tick15/.initial=brown,tick17/.initial=brown,tick19/.initial=brown,tick21/.initial=brown,tick23/.initial=brown,
yticklabel style={color=\pgfkeysvalueof{/pgfplots/tick\ticknum},
},
},
xlabel = Years, 
xbar stacked,
xticklabel style = {/pgf/number format/1000 sep=},
xmax = 2013,
bar width=6pt]

\addplot [white] coordinates {
(2003,Djokovic)
(2001,Nadal)
(2000,Roddick)
(1998,Federer)
(1998,Ferrero)
(1998,Hewitt)
(1997,Safin)
(1995,Kuerten)
(1995,Moya)
(1994,Rios)
(1992,Kafelnikov)
(1991,Rafter)
(1988,Courier)
(1988,Sampras)
(1986,Agassi)
(1985,Muster)
(1984,Becker)
(1983,Edberg)
(1981,Wilander)
(1978,Lendl)
(1978,McEnroe)
(1973,Borg)
(1972,Connors)
(1969,Nastase)
(1968,Newcombe)
};

\addplot [blue,fill=blue] coordinates {
(10,Djokovic)
(0,Nadal)
(12,Roddick)
(0,Federer)
(14,Ferrero)
(0,Hewitt)
(12,Safin)
(0,Kuerten)
(15,Moya)
(0,Rios)
(11,Kafelnikov)
(0,Rafter)
(12,Courier)
(0,Sampras)
(20,Agassi)
(0,Muster)
(15,Becker)
(0,Edberg)
(15,Wilander)
(0,Lendl)
(14,McEnroe)
(0,Borg)
(24,Connors)
(0,Nastase)
(13,Newcombe)
};

\addplot [brown,fill=brown] coordinates {
(0,Djokovic)
(12,Nadal)
(0,Roddick)
(15,Federer)
(0,Ferrero)
(15,Hewitt)
(0,Safin)
(13,Kuerten)
(0,Moya)
(10,Rios)
(0,Kafelnikov)
(11,Rafter)
(0,Courier)
(14,Sampras)
(0,Agassi)
(14,Muster)
(0,Becker)
(13,Edberg)
(0,Wilander)
(16,Lendl)
(0,McEnroe)
(10,Borg)
(0,Connors)
(16,Nastase)
(0,Newcombe)
};

\addplot [blue,fill=white] coordinates {
(10,Djokovic)
(0,Nadal)
(1,Roddick)
(0,Federer)
(1,Ferrero)
(0,Hewitt)
(4,Safin)
(0,Kuerten)
(3,Moya)
(0,Rios)
(10,Kafelnikov)
(0,Rafter)
(13,Courier)
(0,Sampras)
(7,Agassi)
(0,Muster)
(14,Becker)
(0,Edberg)
(17,Wilander)
(0,Lendl)
(21,McEnroe)
(0,Borg)
(17,Connors)
(0,Nastase)
(32,Newcombe)
};

\addplot [brown,fill=white] coordinates {
(0,Djokovic)
(0,Nadal)
(0,Roddick)
(0,Federer)
(0,Ferrero)
(0,Hewitt)
(0,Safin)
(5,Kuerten)
(0,Moya)
(9,Rios)
(0,Kafelnikov)
(11,Rafter)
(0,Courier)
(11,Sampras)
(0,Agassi)
(14,Muster)
(0,Becker)
(17,Edberg)
(0,Wilander)
(19,Lendl)
(0,McEnroe)
(30,Borg)
(0,Connors)
(28,Nastase)
(0,Newcombe)
};
\end{axis}
\end{tikzpicture}
\end{figure}
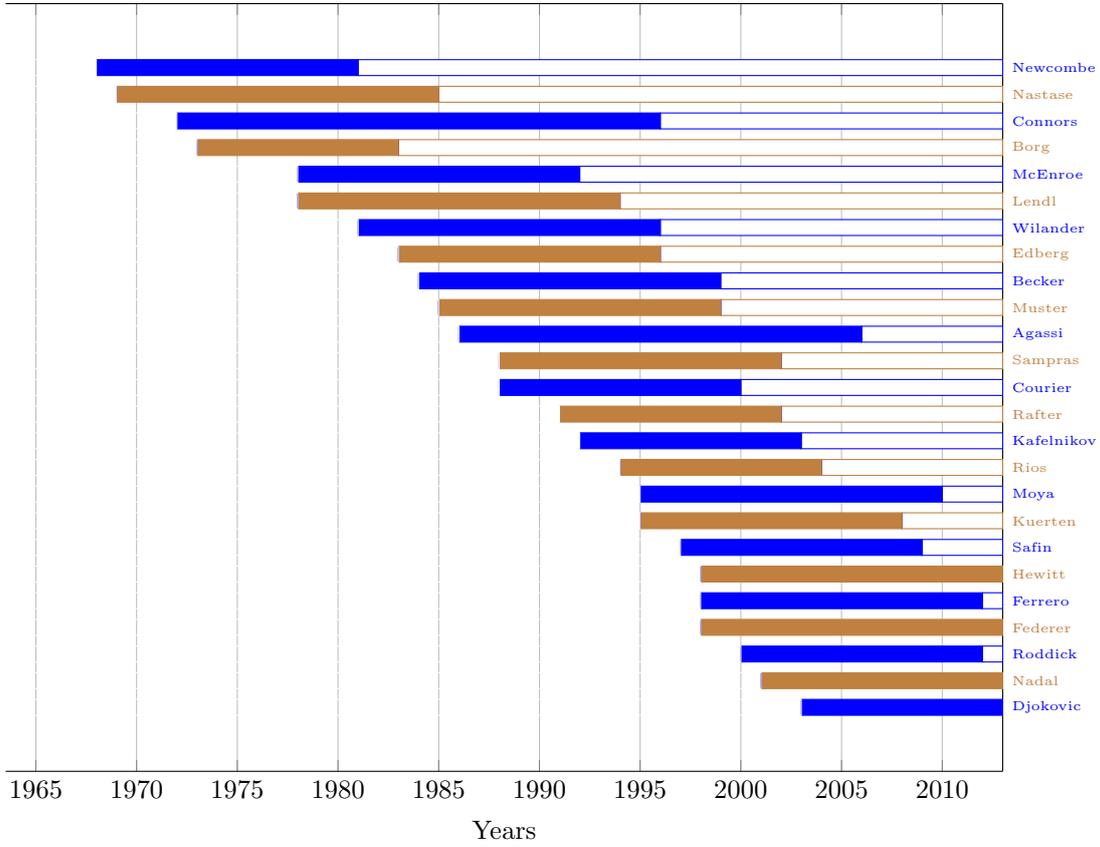

We have chosen those 25 players who have been \#1 on the ATP ranking for any period of time from 1973. 
Figure~\ref{Fig1} shows them together with their active period in the world of professional tennis.

In our calculations the initial data are as follows:
\begin{itemize}
\item
$z_{ij} \, (i, j = 1, \dots ,n, \, i \neq j)$: the number of matches have been played between players $P_i$ and $P_j$ ($z_{ij} = z_{ji}$);
\item
$x_{ij} \, (i > j)$: the number of matches between players $P_i$ and $P_j$, where $P_i$ was the winner;
\item
$y_{ij} = z_{ij} - x_{ij} \, (i > j)$: the number of matches between players $P_i$ and $P_j$, where $P_i$ lost against $P_j$.
\end{itemize}

\begin{definition} \label{Def24}
\textbf{Pairwise comparison matrix of top tennis players}:
$p_{ij}$ elements of matrix $P$ are calculated from the initial data as 
\begin{itemize}
\item
$p_{ij} = x_{ij} / y_{ij}$ if $i, j = 1, \dots ,n$, $i > j$ and $x_{ij} \neq 0$, $y_{ij} \neq 0$;
\item
$p_{ji} = y_{ij} / x_{ij} = 1 / p_{ij}$ if $i, j = 1, \dots ,n$, $i < j$ and $x_{ij} \neq 0$, $z_{ij} \neq 0$;
\item
$p_{ii} = 1$ for all $i = 1, \dots ,n$;
\item
$p_{ij}$ and $p_{ji}$ elements are missing otherwise.
\end{itemize}
\end{definition}

A consequence of the definition is that in case of $z_{ij} = 0$ for at least one pair of the players, the pairwise comparison matrix is incomplete. The interpretation of $p_{ij} > 0$ is that the $i$th player is $p_{ij}$ times better than the $j$th player. 

We have to note that Definition~\ref{Def24} is strict in the sense that $p_{ij}$ is also missing in the case when $z_{ij} \neq 0$, but one of its component is $0$ (either $x_{ij} = 0$ or $y_{ij} = 0$). However, it can happen that $P_i$ won several times over $P_j$, and he has never been defeated. According to the definition we can eliminate all pairs where that phenomenon occurs, but it would be unfair for the winner player in the given pair. Therefore we decided to use artificial $p_{ij}$ values for these cases. For instance $p_{ij} = 5$ was used if $z_{ij}$ was less than $5$ and $y_{ij} = 0$, $p_{ij} = 10$ if the number of matches was between $6$ and $10$, and $y_{ij} = 0$, and so on. In our calculations we will use this correction method and we will refer to it with a subscript $1$. Another correction method for $p_{ij}$ could be that the value of $p_{ij} = x_{ij} + 2$ if $y_{ij} = 0$. In our calculations we will refer to that correction method with a subscript $2$ (see Table \ref{Table2} and \ref{Table3} later).

One can naturally argue that the choice of $p_{ij}$ is crucial to get different results. We have made a series of calculations for various numbers of players with several correction values \citep{TemesiCsatoBozoki2012} and we have found that the results did not alter significantly.

Table~\ref{Table1} contains the results of the matches played between players $P_i$ and $P_j$ (sum of the symmetric elements of the matrix is $z_{ij}$ for all players). We can see that there are cases when two other players have played more than $30$ times with each other, and it was also possible that two players met less than $5$ times. There is a need for balancing the impact of extremely differing match numbers resulted in a wide range of ratios. In order to handle that problem we introduced a transformation for the elements of $p_{ij}$:
\begin{equation} \label{eq7}
t_{ij} = p_{ij}^{z_{ij} / \max z_{ij}}
\end{equation}
where the transforming factor is the ratio of the number of matches between each other divided by the maximum number of matches of all pairs.

Note that if all players have the same number of matches, transformation \eqref{eq7} results in $t_{ij} = p_{ij}$. It approximates $1$ when the two players have played a small number of matches against each other, therefore the outcome seems to be 'unreliable'.
For instance, the original $p_{ij}$ value for the pair \emph{Agassi}-\emph{Becker} was $10/4 = 2.5$, the transformed $t_{ij}$ value is $(10/4)^{14/39} = 1.3895$ where $14$ is the number of matches between \emph{Agassi} and \emph{Becker}, and $39$ is the maximum of $z_{ij}$ values (\emph{Djokovic} vs. \emph{Nadal}).

The vertices of the graph in Figure~\ref{Fig2} represent the players. The edges show that the two players played at least one match against each other. The bold lines connected to the node labelled \emph{Agassi} illustrate that he played against $20$ of our players during his carrier: the degree of the vertex is $20$ (which is also the maximum degree). Edges from \emph{Agassi} to other players (e.g. to \emph{Connors}) mean that \emph{Agassi} has more wins than losses against them (indicated also in the neighbouring table). Similarly, edges to \emph{Agassi} from other players (e.g. from \emph{Rios}) mean that \emph{Agassi} has more losses than wins against them, while dashed lines represent an equal number of wins and losses (e.g. to \emph{Safin}). We have plotted the graphs belonging to all players in an Online Appendix, available at \url{http://www.sztaki.mta.hu/~bozoki/tennis/appendix.pdf}.

Having the incomplete pairwise comparison matrices for the $25$ top tennis players from Table~\ref{Table1} we can calculate the weight vectors if the corresponding matrix $\mathbf{T}$ is connected.
It can be checked that this condition is met: the 20 edges adjacent to the node \emph{Agassi} together with the edges \emph{Nadal}-\emph{Djokovic}, \emph{Newcombe}-\emph{Nastase}, \emph{Nastase}-\emph{Connors} and \emph{Connors}-\emph{Borg} form a spanning tree (see Figure~\ref{Fig2}).

\newgeometry{top=5mm,bottom=15mm}
\afterpage{%
    \clearpage
\begin{landscape}

\linespread{2}
\begin{table}[htbp]
\caption{Pairwise comparisons: number of wins/total number of matches}
\label{Table1}
\centering
\begin{tiny}
\noindent\makebox[\linewidth]{
\rowcolors{1}{gray!10}{}
    \begin{tabular}{l ccccc ccccc ccccc ccccc ccccc >{\bfseries}c} \toprule \hiderowcolors
          & \begin{sideways}Agassi\end{sideways} & \begin{sideways}Becker\end{sideways} & \begin{sideways}Borg\end{sideways} & \begin{sideways}Connors\end{sideways} & \begin{sideways}Courier\end{sideways} & \begin{sideways}Djokovic\end{sideways} & \begin{sideways}Edberg\end{sideways} & \begin{sideways}Federer\end{sideways} & \begin{sideways}Ferrero\end{sideways} & \begin{sideways}Hewitt\end{sideways} & \begin{sideways}Kafelnikov\end{sideways} & \begin{sideways}Kuerten\end{sideways} & \begin{sideways}Lendl\end{sideways} & \begin{sideways}McEnroe\end{sideways} & \begin{sideways}Moya\end{sideways} & \begin{sideways}Muster\end{sideways} & \begin{sideways}Nadal\end{sideways} & \begin{sideways}Nastase\end{sideways} & \begin{sideways}Newcombe\end{sideways} & \begin{sideways}Rafter\end{sideways} & \begin{sideways}Rios\end{sideways} & \begin{sideways}Roddick\end{sideways} & \begin{sideways}Safin\end{sideways} & \begin{sideways}Sampras\end{sideways} & \begin{sideways}Wilander\end{sideways} & Total\\ \midrule \showrowcolors
    Agassi & \cellcolor{gray!80} & \textcolor{green}{10/14} &       & \textcolor{green}{2/2} & \textcolor{red}{5/12} &       & \textcolor{green}{6/9} & \textcolor{red}{3/11} & \textcolor{red}{2/5} & \textcolor{blue}{4/8} & \textcolor{green}{8/12} & \textcolor{green}{7/11} & \textcolor{red}{2/8} & \textcolor{blue}{2/4} & \textcolor{green}{3/4} & \textcolor{green}{5/9} & \textcolor{red}{0/2} &       &       & \textcolor{green}{10/15} & \textcolor{red}{1/3} & \textcolor{green}{5/6} & \textcolor{blue}{3/6} & \textcolor{red}{14/34} & \textcolor{green}{5/7} & \textcolor{green}{97/182} \\
    Becker & \textcolor{red}{4/14} & \cellcolor{gray!80} &       & \textcolor{green}{6/6} & \textcolor{green}{6/7} &       & \textcolor{green}{25/35} &       &       & \textcolor{green}{1/1} & \textcolor{green}{4/6} &       & \textcolor{red}{10/21} & \textcolor{green}{8/10} & \textcolor{blue}{2/4} & \textcolor{green}{2/3} &       & \textcolor{green}{1/1} &       & \textcolor{green}{2/3} & \textcolor{green}{3/5} &       & \textcolor{red}{0/1} & \textcolor{red}{7/19} & \textcolor{green}{7/10} & \textcolor{green}{88/146} \\
    Borg  &       &       & \cellcolor{gray!80} & \textcolor{green}{15/23} &       &       &       &       &       &       &       &       & \textcolor{green}{6/8} & \textcolor{blue}{7/14} &       &       &       & \textcolor{green}{10/15} & \textcolor{red}{1/4} &       &       &       &       &       & \textcolor{green}{1/1} & \textcolor{green}{40/65} \\
    Connors & \textcolor{red}{0/2} & \textcolor{red}{0/6} & \textcolor{red}{8/23} & \cellcolor{gray!80} & \textcolor{red}{0/3} &       & \textcolor{blue}{6/12} &       &       &       &       &       & \textcolor{red}{13/34} & \textcolor{red}{14/34} &       &       &       & \textcolor{red}{12/27} & \textcolor{blue}{2/4} &       &       &       &       & \textcolor{red}{0/2} & \textcolor{red}{0/5} & \textcolor{red}{55/152} \\
    Courier & \textcolor{green}{7/12} & \textcolor{red}{1/7} &       & \textcolor{green}{3/3} & \cellcolor{gray!80} &       & \textcolor{green}{6/10} &       &       &       & \textcolor{red}{1/6} & \textcolor{green}{1/1} & \textcolor{red}{0/4} & \textcolor{green}{2/3} & \textcolor{green}{2/3} & \textcolor{green}{7/12} &       & \textcolor{red}{0/1} &       & \textcolor{red}{0/3} & \textcolor{red}{0/3} &       & \textcolor{blue}{1/2} & \textcolor{red}{4/20} &       & \textcolor{red}{35/90} \\
    Djokovic &       &       &       &       &       & \cellcolor{gray!80} &       & \textcolor{red}{15/31} & \textcolor{green}{2/3} & \textcolor{green}{6/7} &       &       &       &       & \textcolor{blue}{2/4} &       & \textcolor{red}{17/39} &       &       &       &       & \textcolor{red}{4/9} & \textcolor{red}{0/2} &       &       & \textcolor{red}{46/95} \\
    Edberg & \textcolor{red}{3/9} & \textcolor{red}{10/35} &       & \textcolor{blue}{6/12} & \textcolor{red}{4/10} &       & \cellcolor{gray!80} &       &       &       & \textcolor{red}{1/3} &       & \textcolor{green}{14/27} & \textcolor{red}{6/13} & \textcolor{green}{1/1} & \textcolor{green}{10/10} &       &       &       & \textcolor{green}{3/3} & \textcolor{green}{1/1} &       &       & \textcolor{red}{6/14} & \textcolor{red}{9/20} & \textcolor{red}{74/158} \\
    Federer & \textcolor{green}{8/11} &       &       &       &       & \textcolor{green}{16/31} &       & \cellcolor{gray!80} & \textcolor{green}{10/13} & \textcolor{green}{18/26} & \textcolor{red}{2/6} & \textcolor{red}{1/3} &       &       & \textcolor{green}{7/7} &       & \textcolor{red}{10/32} &       &       & \textcolor{red}{0/3} & \textcolor{green}{2/2} & \textcolor{green}{21/24} & \textcolor{green}{10/12} & \textcolor{green}{1/1} &       & \textcolor{green}{106/171} \\
    Ferrero & \textcolor{green}{3/5} &       &       &       &       & \textcolor{red}{1/3} &       & \textcolor{red}{3/13} & \cellcolor{gray!80} & \textcolor{red}{4/10} & \textcolor{red}{1/3} & \textcolor{green}{3/5} &       &       & \textcolor{green}{8/14} &       & \textcolor{red}{2/9} &       &       & \textcolor{green}{2/3} & \textcolor{green}{3/4} & \textcolor{red}{0/5} & \textcolor{blue}{6/12} &       &       & \textcolor{red}{36/86} \\
    Hewitt & \textcolor{blue}{4/8} & \textcolor{red}{0/1} &       &       &       & \textcolor{red}{1/7} &       & \textcolor{red}{8/26} & \textcolor{green}{6/10} & \cellcolor{gray!80} & \textcolor{green}{7/8} & \textcolor{green}{3/4} &       &       & \textcolor{green}{7/12} &       & \textcolor{red}{4/10} &       &       & \textcolor{green}{3/4} & \textcolor{green}{3/5} & \textcolor{blue}{7/14} & \textcolor{blue}{7/14} & \textcolor{green}{5/9} &       & \textcolor{red}{65/132} \\
    Kafelnikov & \textcolor{red}{4/12} & \textcolor{red}{2/6} &       &       & \textcolor{green}{5/6} &       & \textcolor{green}{2/3} & \textcolor{green}{4/6} & \textcolor{green}{2/3} & \textcolor{red}{1/8} & \cellcolor{gray!80} & \textcolor{red}{5/12} &       &       & \textcolor{blue}{3/6} & \textcolor{red}{1/5} &       &       &       & \textcolor{green}{3/5} & \textcolor{green}{6/8} &       & \textcolor{blue}{2/4} & \textcolor{red}{2/13} & \textcolor{blue}{1/2} & \textcolor{red}{43/99} \\
    Kuerten & \textcolor{red}{4/11} &       &       &       & \textcolor{red}{0/1} &       &       & \textcolor{green}{2/3} & \textcolor{red}{2/5} & \textcolor{red}{1/4} & \textcolor{green}{7/12} & \cellcolor{gray!80} &       &       & \textcolor{green}{4/7} & \textcolor{green}{3/3} &       &       &       & \textcolor{blue}{4/8} & \textcolor{blue}{2/4} & \textcolor{blue}{1/2} & \textcolor{green}{4/7} & \textcolor{red}{1/3} &       & \textcolor{blue}{35/70} \\
    Lendl & \textcolor{green}{6/8} & \textcolor{green}{11/21} & \textcolor{red}{2/8} & \textcolor{green}{21/34} & \textcolor{green}{4/4} &       & \textcolor{red}{13/27} &       &       &       &       &       & \cellcolor{gray!80} & \textcolor{green}{21/36} &       & \textcolor{green}{4/5} &       & \textcolor{green}{1/1} &       & \textcolor{red}{0/1} &       &       &       & \textcolor{red}{3/8} & \textcolor{green}{15/22} & \textcolor{green}{101/175} \\
    McEnroe & \textcolor{blue}{2/4} & \textcolor{red}{2/10} & \textcolor{blue}{7/14} & \textcolor{green}{20/34} & \textcolor{red}{1/3} &       & \textcolor{green}{7/13} &       &       &       &       &       & \textcolor{red}{15/36} & \cellcolor{gray!80} &       &       &       & \textcolor{green}{6/9} & \textcolor{blue}{1/2} &       &       &       &       & \textcolor{red}{0/3} & \textcolor{green}{7/13} & \textcolor{red}{68/141} \\
    Moya  & \textcolor{red}{1/4} & \textcolor{blue}{2/4} &       &       & \textcolor{red}{1/3} & \textcolor{blue}{2/4} & \textcolor{red}{0/1} & \textcolor{red}{0/7} & \textcolor{red}{6/14} & \textcolor{red}{5/12} & \textcolor{blue}{3/6} & \textcolor{red}{3/7} &       &       & \cellcolor{gray!80} & \textcolor{blue}{4/8} & \textcolor{red}{2/8} &       &       & \textcolor{green}{3/4} & \textcolor{red}{2/7} & \textcolor{red}{1/5} & \textcolor{green}{4/7} & \textcolor{red}{1/4} &       & \textcolor{red}{40/105} \\
    Muster & \textcolor{red}{4/9} & \textcolor{red}{1/3} &       &       & \textcolor{red}{5/12} &       & \textcolor{red}{0/10} &       &       &       & \textcolor{green}{4/5} & \textcolor{red}{0/3} & \textcolor{red}{1/5} &       & \textcolor{blue}{4/8} & \cellcolor{gray!80} &       &       &       & \textcolor{red}{0/3} & \textcolor{green}{3/4} &       & \textcolor{red}{0/1} & \textcolor{red}{2/11} & \textcolor{red}{0/2} & \textcolor{red}{24/76} \\
    Nadal & \textcolor{green}{2/2} &       &       &       &       & \textcolor{green}{22/39} &       & \textcolor{green}{22/32} & \textcolor{green}{7/9} & \textcolor{green}{6/10} &       &       &       &       & \textcolor{green}{6/8} &       & \cellcolor{gray!80} &       &       &       &       & \textcolor{green}{7/10} & \textcolor{green}{2/2} &       &       & \textcolor{green}{74/112} \\
    Nastase &       & \textcolor{red}{0/1} & \textcolor{red}{5/15} & \textcolor{green}{15/27} & \textcolor{green}{1/1} &       &       &       &       &       &       &       & \textcolor{red}{0/1} & \textcolor{red}{3/9} &       &       &       & \cellcolor{gray!80} & \textcolor{green}{4/5} &       &       &       &       &       & \textcolor{red}{0/1} & \textcolor{red}{28/60} \\
    Newcombe &       &       & \textcolor{green}{3/4} & \textcolor{blue}{2/4} &       &       &       &       &       &       &       &       &       & \textcolor{blue}{1/2} &       &       &       & \textcolor{red}{1/5} & \cellcolor{gray!80} &       &       &       &       &       &       & \textcolor{red}{7/15} \\
    Rafter & \textcolor{red}{5/15} & \textcolor{red}{1/3} &       &       & \textcolor{green}{3/3} &       & \textcolor{red}{0/3} & \textcolor{green}{3/3} & \textcolor{red}{1/3} & \textcolor{red}{1/4} & \textcolor{red}{2/5} & \textcolor{blue}{4/8} & \textcolor{green}{1/1} &       & \textcolor{red}{1/4} & \textcolor{green}{3/3} &       &       &       & \cellcolor{gray!80} & \textcolor{green}{2/3} &       & \textcolor{green}{1/1} & \textcolor{red}{4/16} & \textcolor{red}{1/3} & \textcolor{red}{33/78} \\
    Rios  & \textcolor{green}{2/3} & \textcolor{red}{2/5} &       &       & \textcolor{green}{3/3} &       & \textcolor{red}{0/1} & \textcolor{red}{0/2} & \textcolor{red}{1/4} & \textcolor{red}{2/5} & \textcolor{red}{2/8} & \textcolor{blue}{2/4} &       &       & \textcolor{green}{5/7} & \textcolor{red}{1/4} &       &       &       & \textcolor{red}{1/3} & \cellcolor{gray!80} & \textcolor{red}{0/2} & \textcolor{red}{1/4} & \textcolor{red}{0/2} &       & \textcolor{red}{22/57} \\
    Roddick & \textcolor{red}{1/6} &       &       &       &       & \textcolor{green}{5/9} &       & \textcolor{red}{3/24} & \textcolor{green}{5/5} & \textcolor{blue}{7/14} &       & \textcolor{blue}{1/2} &       &       & \textcolor{green}{4/5} &       & \textcolor{red}{3/10} &       &       &       & \textcolor{green}{2/2} & \cellcolor{gray!80} & \textcolor{green}{4/7} & \textcolor{green}{2/3} &       & \textcolor{red}{37/87} \\
    Safin & \textcolor{blue}{3/6} & \textcolor{green}{1/1} &       &       & \textcolor{blue}{1/2} & \textcolor{green}{2/2} &       & \textcolor{red}{2/12} & \textcolor{blue}{6/12} & \textcolor{blue}{7/14} & \textcolor{blue}{2/4} & \textcolor{red}{3/7} &       &       & \textcolor{red}{3/7} & \textcolor{green}{1/1} & \textcolor{red}{0/2} &       &       & \textcolor{red}{0/1} & \textcolor{green}{3/4} & \textcolor{red}{3/7} & \cellcolor{gray!80} & \textcolor{green}{4/7} &       & \textcolor{red}{41/89} \\
    Sampras & \textcolor{green}{20/34} & \textcolor{green}{12/19} &       & \textcolor{green}{2/2} & \textcolor{green}{16/20} &       & \textcolor{green}{8/14} & \textcolor{red}{0/1} &       & \textcolor{red}{4/9} & \textcolor{green}{11/13} & \textcolor{green}{2/3} & \textcolor{green}{5/8} & \textcolor{green}{3/3} & \textcolor{green}{3/4} & \textcolor{green}{9/11} &       &       &       & \textcolor{green}{12/16} & \textcolor{green}{2/2} & \textcolor{red}{1/3} & \textcolor{red}{3/7} & \cellcolor{gray!80} & \textcolor{green}{2/3} & \textcolor{green}{115/172} \\
    Wilander & \textcolor{red}{2/7} & \textcolor{red}{3/10} & \textcolor{red}{0/1} & \textcolor{green}{5/5} &       &       & \textcolor{green}{11/20} &       &       &       & \textcolor{blue}{1/2} &       & \textcolor{red}{7/22} & \textcolor{red}{6/13} &       & \textcolor{green}{2/2} &       & \textcolor{green}{1/1} &       & \textcolor{green}{2/3} &       &       &       & \textcolor{red}{1/3} & \cellcolor{gray!80} & \textcolor{red}{41/89} \\ \bottomrule
    \end{tabular} }
\end{tiny}
\end{table} 

\linespread{1}
\end{landscape}
    \clearpage
}
\restoregeometry

\newgeometry{top=0mm,bottom=5mm,left=0mm,right=0mm}
\begin{landscape}
\pagestyle{empty}

\tikzset{->-/.style={decoration={
  markings,
  mark=at position .5 with {\arrow{>}}},postaction={decorate}}}

\begin{figure}[htbp]
\caption{Graph representation of matrix $\mathbf{T}$}
\label{Fig2}
\begin{subfigure}{1.08\textheight}
\begin{tikzpicture}[scale=0.93, auto=center, transform shape,>=triangle 45]
  \node[circle,fill=black] (n1) at (6*360/25:10) {};
      \node (n1a) at (6*360/25:10.5) {Newcombe};
  \node[circle,fill=black] (n2) at (5*360/25:10) {};
    \node (n2a) at (5*360/25:10.6) {Nastase};
  \node[circle,fill=black] (n3) at (4*360/25:10) {};
    \node (n3a) at (4*360/25:10.7) {Connors};
  \node[circle,fill=black] (n4) at (3*360/25:10) {};
    \node (n4a) at (3*360/25:10.8) {Borg};
  \node[circle,fill=black] (n5) at (2*360/25:10) {};
    \node (n5a) at (2*360/25:10.9) {McEnroe};
  \node[circle,fill=black] (n6) at (1*360/25:10) {};
    \node (n6a) at (1*360/25:11) {Lendl};
  \node[circle,fill=black] (n7) at (0*360/25:10) {};
    \node (n7a) at (0*360/25:11.1) {Wilander};
  \node[circle,fill=black] (n8) at (24*360/25:10) {};
    \node (n8a) at (24*360/25:11) {Edberg};
  \node[circle,fill=black] (n9) at (23*360/25:10) {};
    \node (n9a) at (23*360/25:10.9) {Becker};
  \node[circle,fill=black] (n10) at (22*360/25:10) {};
    \node (n10a) at (22*360/25:10.8) {Muster};
  \node[circle,fill=black] (n11) at (21*360/25:10) {};
    \node (n11a) at (21*360/25:10.7) {Agassi};
  \node[circle,fill=black] (n12) at (20*360/25:10) {};
    \node (n12a) at (20*360/25:10.6) {Courier};
  \node[circle,fill=black] (n13) at (19*360/25:10) {};
    \node (n13a) at (19*360/25:10.5) {Sampras};
  \node[circle,fill=black] (n14) at (18*360/25:10) {};
    \node (n14a) at (18*360/25:10.6) {Rafter};
  \node[circle,fill=black] (n15) at (17*360/25:10) {};
    \node (n15a) at (17*360/25:10.7) {Kafelnikov};
  \node[circle,fill=black] (n16) at (16*360/25:10) {};
    \node (n16a) at (16*360/25:10.8) {Rios};
  \node[circle,fill=black] (n17) at (15*360/25:10) {};
    \node (n17a) at (15*360/25:10.9) {Moya};
  \node[circle,fill=black] (n18) at (14*360/25:10) {};
    \node (n18a) at (14*360/25:11) {Kuerten};
  \node[circle,fill=black] (n19) at (13*360/25:10) {};
    \node (n19a) at (13*360/25:11.1) {Safin};
  \node[circle,fill=black] (n20) at (12*360/25:10) {};
    \node (n20a) at (12*360/25:11) {Ferrero};
  \node[circle,fill=black] (n21) at (11*360/25:10) {};
    \node (n21a) at (11*360/25:10.9) {Hewitt};
  \node[circle,fill=black] (n22) at (10*360/25:10) {};
    \node (n22a) at (10*360/25:10.8) {Federer};
  \node[circle,fill=black] (n23) at (9*360/25:10) {};
    \node (n23a) at (9*360/25:10.7) {Roddick};
  \node[circle,fill=black] (n24) at (8*360/25:10) {};
    \node (n24a) at (8*360/25:10.6) {Nadal};
  \node[circle,fill=black] (n25) at (7*360/25:10) {};
    \node (n25a) at (7*360/25:10.5) {Djokovic};

  \foreach \from/\to in {n1/n2,n1/n3,n1/n5,n2/n3,n2/n5,n2/n6,n2/n7,n2/n9,n2/n12,n3/n4,n3/n5,n3/n6,n3/n7,n3/n8,n3/n9,n3/n11,n3/n12,n3/n13,n4/n5,n4/n6,n4/n7,n5/n6,n5/n7,n5/n8,n5/n9,n5/n11,n5/n12,n5/n13,n6/n7,n6/n8,n6/n9,n6/n10,n6/n11,n6/n12,n6/n13,n6/n14,n7/n8,n7/n9,n7/n10,n7/n11,n7/n13,n7/n14,n7/n15,n8/n9,n8/n10,n8/n11,n8/n12,n8/n13,n8/n14,n8/n15,n8/n16,n8/n17,n9/n10,n9/n11,n9/n12,n9/n13,n9/n14,n9/n15,n9/n16,n9/n17,n9/n19,n9/n21,n10/n11,n10/n12,n10/n13,n10/n14,n10/n15,n10/n16,n10/n17,n10/n18,n10/n19,n11/n12,n11/n13,n11/n14,n11/n15,n11/n16,n11/n17,n11/n18,n11/n19,n11/n20,n11/n21,n11/n22,n11/n23,n11/n24,n12/n13,n12/n14,n12/n15,n12/n16,n12/n17,n12/n18,n12/n19,n13/n14,n13/n15,n13/n16,n13/n17,n13/n18,n13/n19,n13/n21,n13/n22,n13/n23,n14/n15,n14/n16,n14/n17,n14/n18,n14/n19,n14/n20,n14/n21,n14/n22,n15/n16,n15/n17,n15/n18,n15/n19,n15/n20,n15/n21,n15/n22,n16/n17,n16/n18,n16/n19,n16/n20,n16/n21,n16/n22,n16/n23,n17/n18,n17/n19,n17/n20,n17/n21,n17/n22,n17/n23,n17/n24,n17/n25,n18/n19,n18/n20,n18/n21,n18/n22,n18/n23,n19/n20,n19/n21,n19/n22,n19/n23,n19/n24,n19/n25,n20/n21,n20/n22,n20/n23,n20/n24,n20/n25,n21/n22,n21/n23,n21/n24,n21/n25,n22/n23,n22/n24,n22/n25,n23/n24,n23/n25,n24/n25}
    \draw[color=lightgray] (\from) -- (\to);
  \foreach \from/\to in {n6/n11,n12/n11,n13/n11,n16/n11,n20/n11,n22/n11,n24/n11}
    \draw[color=red,->-] (\from) -- (\to);
  \foreach \from/\to in {n11/n3,n11/n7,n11/n8,n11/n9,n11/n10,n11/n14,n11/n15,n11/n17,n11/n18,n11/n23}
    \draw[color=green,->-] (\from) -- (\to);
  \foreach \from/\to in {n5/n11,n19/n11,n21/n11}
    \draw[color=white] (\from) -- (\to);
  \foreach \from/\to in {n5/n11,n19/n11,n21/n11}
    \draw[color=blue,dashed,thick] (\from) -- (\to);
\end{tikzpicture}
\end{subfigure}
\begin{subfigure}{.3\textheight}
\begin{tabularx}{.25\textheight}{Lcc}
    \multicolumn{3}{c}{\textbf{Agassi}} \\ \midrule
    against & Win   & Loss \\ \midrule
    Becker & 10    & 4 \\
    Connors & 2     & 0 \\
    Courier & 5     & 7 \\
    Edberg & 6     & 3 \\
    Federer & 3     & 8 \\
    Ferrero & 2     & 3 \\
    Hewitt & 4     & 4 \\
    Kafelnikov & 8     & 4 \\
    Kuerten & 7     & 4 \\
    Lendl & 2     & 6 \\
    McEnroe & 2     & 2 \\
    Moya  & 3     & 1 \\
    Muster & 5     & 4 \\
    Nadal & 0     & 2 \\
    Rafter & 10    & 5 \\
    Rios  & 1     & 2 \\
    Roddick & 5     & 1 \\
    Safin & 3     & 3 \\
    Sampras & 14    & 20 \\
    Wilander & 5     & 2 \\ \midrule
    Sum   & 97    & 85 \\ \bottomrule
\end{tabularx}
\end{subfigure}
\end{figure}
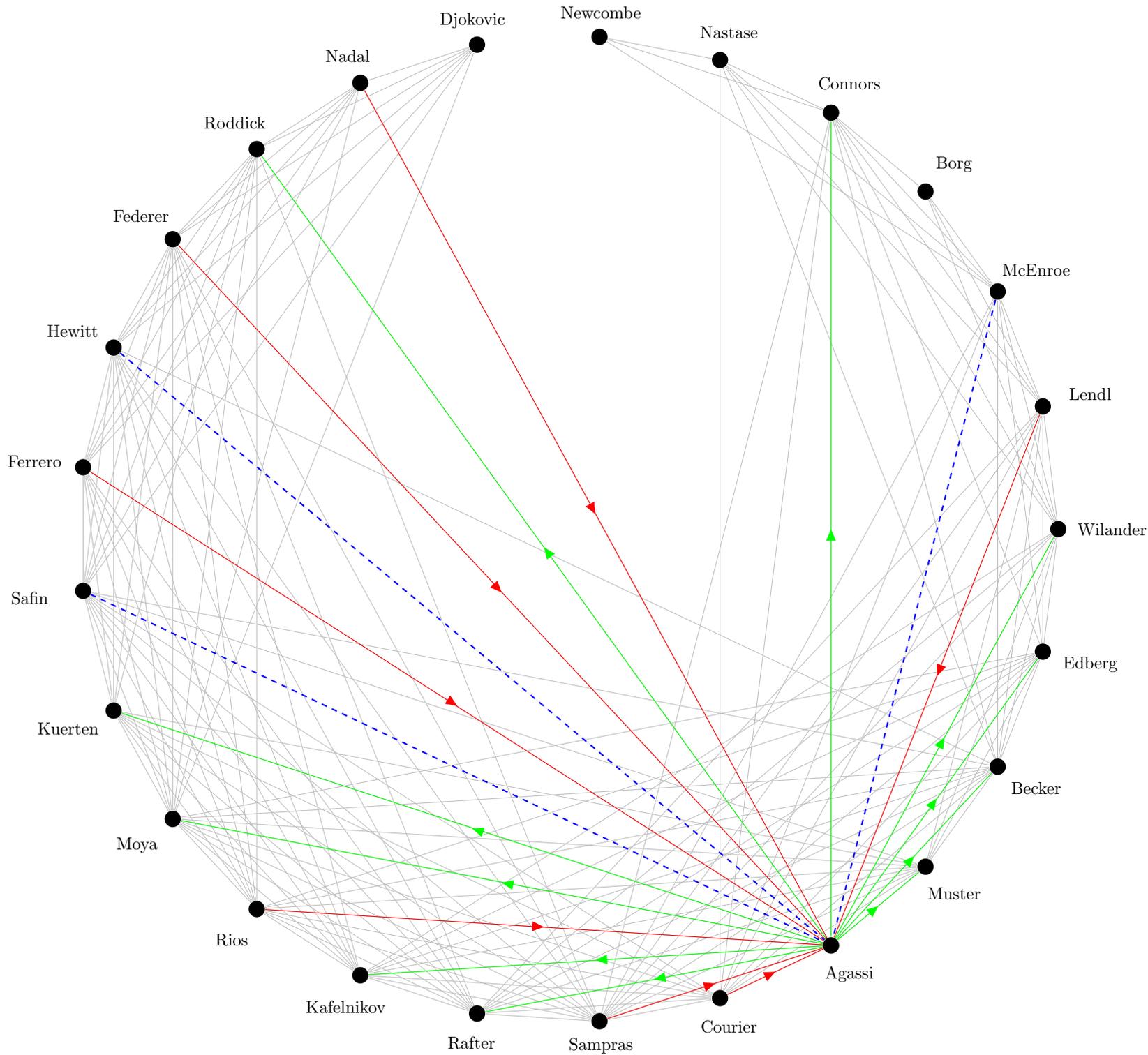  

\end{landscape}
\restoregeometry

\begin{table}[h]
  \centering
  \caption{Rankings}
  \label{Table2}
\rowcolors{1}{gray!30}{}
    \begin{tabularx}{\textwidth}{l CCC>{\bfseries}CC} \toprule \hiderowcolors
     & $EM_2$  & $LLSM_2$ & $EM_{W2}$ & $LLSM_{W2}$ & W / L \\ \midrule \showrowcolors
    Nadal & 1     & 1     & 1     & 1     & 2 \\
    Federer & 2     & 2     & 2     & 2     & 3 \\
    Sampras & 3     & 3     & 3     & 3     & 1 \\
    Lendl & 11    & 8     & 4     & 4     & 6 \\
    Borg  & 13    & 11    & 6     & 5     & 4 \\
    Becker & 4     & 4     & 5     & 6     & 5 \\
    Djokovic & 5     & 5     & 7     & 7     & 10 \\
    Agassi & 9     & 9     & 8     & 8     & 7 \\
    Hewitt & 6     & 7     & 9     & 9     & 9 \\
    Kuerten & 16    & 15    & 10    & 10    & 8 \\
    Safin & 12    & 10    & 11    & 11    & 15 \\
    McEnroe & 20    & 18    & 12    & 12    & 11 \\
    Nastase & 22    & 20    & 14    & 13    & 13 \\
    Ferrero & 17    & 16    & 16    & 14    & 20 \\
    Roddick & 8     & 6     & 13    & 15    & 18 \\
    Wilander & 15    & 14    & 17    & 16    & 16 \\
    Rios  & 21    & 22    & 18    & 17    & 22 \\
    Rafter & 7     & 13    & 15    & 18    & 19 \\
    Newcombe & 23    & 21    & 21    & 19    & 14 \\
    Kafelnikov & 14    & 17    & 19    & 20    & 17 \\
    Moya  & 19    & 19    & 22    & 21    & 23 \\
    Edberg & 10    & 12    & 20    & 22    & 12 \\
    Courier & 18    & 23    & 23    & 23    & 21 \\
    Muster & 24    & 24    & 24    & 24    & 25 \\
    Connors & 25    & 25    & 25    & 25    & 24 \\ \bottomrule
    \end{tabularx}
\end{table}

Weight vectors have been computed with the Logarithmic Least Squares Method ($LLSM$ in Table~\ref{Table2}) and with the Eigenvector Method ($EM$) as it was described in Section~\ref{Sec2}. On the basis of the weight vectors, eight rankings have been calculated without and with transformation (in the latter case we used the subscript $W$ for identification), and different correction methods have also been applied (subscripts $1$ and $2$ as it was introduced earlier). Selected results are demonstrated in Table~\ref{Table2}. The fourth column, $LLSM_{W2}$, for example, is a ranking given by Logarithmic Least Squares Method with the second correction procedure and transformed data. Note that the players are listed in Table~\ref{Table2} according to this ranking. The fifth column includes the ranking according to the win to loss ratio, indicated by W / L.

\begin{table}[htbp]
\centering
\begin{small}
\caption{Spearman rank correlation coefficients}
\label{Table3}
    \begin{tabular}{l cccc cccc}
    \toprule
     & $EM_1$  & $EM_2$  & $LLSM_1$ & $LLSM_2$ & $EM_{W1}$ & $EM_{W2}$ & $LLSM_{W1}$ & $LLSM_{W2}$ \\
    \midrule
    $EM_1$  & \multicolumn{1}{c}1     & 0.9715 & 0.9269 & 0.9154 & 0.7546 & 0.7423 & 0.6869 & 0.6631 \\
    $EM_2$  & 0.9715 & \multicolumn{1}{c}1     & 0.9677 & 0.9569 & 0.8015 & 0.7908 & 0.7385 & 0.7177 \\
    $LLSM_1$ & 0.9269 & 0.9677 & \multicolumn{1}{c}1     & 0.9915 & 0.8638 & 0.8469 & 0.8085 & 0.7946 \\
    $LLSM_2$ & 0.9154 & 0.9569 & 0.9915 & \multicolumn{1}{c}1     & 0.8931 & 0.8831 & 0.8446 & 0.8338 \\
    $EM_{W1}$ & 0.7546 & 0.8015 & 0.8638 & 0.8931 & \multicolumn{1}{c}1     & 0.9962 & 0.9908 & 0.9854 \\
    $EM_{W2}$ & 0.7423 & 0.7908 & 0.8469 & 0.8831 & 0.9962 & \multicolumn{1}{c}1     & 0.9900 & 0.9877 \\
    $LLSM_{W1}$ & 0.6869 & 0.7385 & 0.8085 & 0.8446 & 0.9908 & 0.9900 & \multicolumn{1}{c}1     & 0.9969 \\
    $LLSM_{W2}$ & 0.6631 & 0.7177 & 0.7946 & 0.8338 & 0.9854 & 0.9877 & 0.9969 & \multicolumn{1}{c}1 \\ \bottomrule
    \end{tabular}
\end{small}
\end{table}

Rankings were practically the same with both estimation methods, as it can be seen from Table~\ref{Table2}. The impact of the correction method is not significant either. (That was the reason why Table~\ref{Table2} does not contains calculations with the first type correction.) The values of the Spearman rank correlation coefficients in Table~\ref{Table3} support these propositions: the elements of the top-left and bottom-right $4 \times 4$ submatrices are close to the identity matrix. The correlation coefficients -- comparing rankings with the same estimation method -- suggest that filtering the impact of differences in the total match numbers eliminated the minor impact of the correction methods, too.
Analysing the impact of the estimation methods and various forms of data correction the authors had similar experience with $34$ top players \citep{TemesiCsatoBozoki2012}.

However, data transformation \eqref{eq7} may change the rankings significantly, as it can be seen in Table~\ref{Table3}, too. The corresponding rank correlation coefficients in the top-right and bottom-left $4 \times 4$ submatrices confirm this statement. According to our interpretation the value judgement of the ranking expert determines the choice between these rankings. Therefore if the expert's opinion is that a ratio of $2$ has to be represented in different ways if it was resulted from $6$ matches ($4:2$) or from $30$ matches ($20:10$) than the recommended normalization has to be implemented and the corresponding ranking can be chosen.

Turning back to Table~\ref{Table2}, the first three players (\emph{Nadal}, \emph{Federer}, \emph{Sampras}) and the last three players (\emph{Courier}, \emph{Muster}, \emph{Connors}) are the same in both rankings. Some differences in the rank numbers can be found in other parts of the list. Tennis fans can debate the final ranking, of course. One can compare these rankings to the win /loss ratio of the players, given in the ninth column of Table~\ref{Table2}. However, the most important fact is that the Top $12$ includes big names from the recent championships and from the good old times, as well. The conclusion is that it is possible to produce rankings based on pairwise comparisons and overarching four decades with players who have never met on the court.

\section{Conclusions and open questions} \label{Sec4}

In case of having historical data incomplete pairwise comparison matrices can be applied in order to answer the question: what is the ranking of the players for a long time period? Who is the \#1 player?
With this methodology it is possible to use face to face match results. Various types of transformations can modify the original data set with the intention of correcting either data problems or biasing factors. However, we did not take into account the impact of  the carrier path of a player. Every match had identical weight without considering its position on the time line. Different surfaces did not play specific role, either.

Having had a great number of calculations with tennis results we have been interested in finding answer for the question 'What are those properties of matrix $\mathbf{T}$ which have an impact on the ranking?'

Ranking can depend on the number and the distribution of the comparisons. The number of comparisons can be characterized by the density (sparsity) of the PCM. For a fully completed PCM the density of the matrix is $1$. Lower values mean that the matrix is incomplete. In our case the density of $\mathbf{T}$ is $341 / 625 \approx 0.5456$ since $341$ elements are known in the $25 \times 25$ matrix.

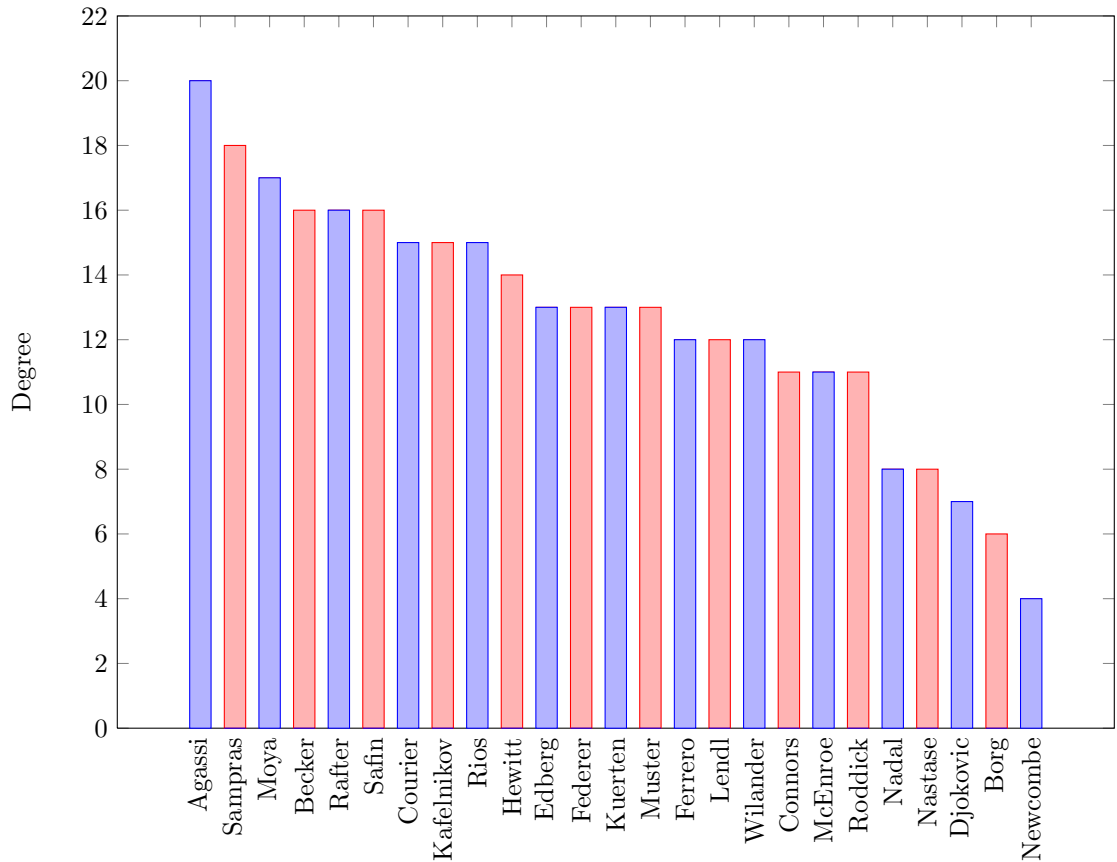
\begin{figure}[h]
\centering
\caption{Degree of vertices in the graph representation of matrix $\mathbf{T}$}
\label{Fig3}

\begin{tikzpicture}
\begin{axis}[width=\textwidth, 
height=0.75\textwidth,
symbolic x coords={Agassi,Sampras,Moya,Becker,Rafter,Safin,Courier,Kafelnikov,Rios,Hewitt,Edberg,Federer,Kuerten,Muster,Ferrero,Lendl,Wilander,Connors,McEnroe,Roddick,Nadal,Nastase,Djokovic,Borg,Newcombe},
xtick=data, 
x tick label style={rotate=90},
ylabel = Degree, 
ybar stacked,
ymin = 0,
bar width=8pt]

\addplot coordinates {
(Agassi,20)
(Sampras,0)
(Moya,17)
(Becker,0)
(Rafter,16)
(Safin,0)
(Courier,15)
(Kafelnikov,0)
(Rios,15)
(Hewitt,0)
(Edberg,13)
(Federer,0)
(Kuerten,13)
(Muster,0)
(Ferrero,12)
(Lendl,0)
(Wilander,12)
(Connors,0)
(McEnroe,11)
(Roddick,0)
(Nadal,8)
(Nastase,0)
(Djokovic,7)
(Borg,0)
(Newcombe,4)
};

\addplot coordinates {
(Agassi,0)
(Sampras,18)
(Moya,0)
(Becker,16)
(Rafter,0)
(Safin,16)
(Courier,0)
(Kafelnikov,15)
(Rios,0)
(Hewitt,14)
(Edberg,0)
(Federer,13)
(Kuerten,0)
(Muster,13)
(Ferrero,0)
(Lendl,12)
(Wilander,0)
(Connors,11)
(McEnroe,0)
(Roddick,11)
(Nadal,0)
(Nastase,8)
(Djokovic,0)
(Borg,6)
(Newcombe,0)
};
\end{axis}
\end{tikzpicture}
\end{figure}

Another indicator of the structure of matrix $\mathbf{T}$ is the distribution of elements, which can be characterized by the degree of vertices in the graph representation of the PCM. We have designed a tool for exploring the connectedness of the incomplete pairwise comparison matrices visually. 
Figure~\ref{Fig3} shows the distribution of degrees in our case, which can be checked in the Online Appendix (\url{http://www.sztaki.mta.hu/~bozoki/tennis/appendix.pdf}), where clicking on a node shows the edges adjacent to it.
The maximum degree is $20$ in the case of \emph{Agassi}.

Moreover, increasing the number of matches played between those who have been played with each other leaves the value of density and the degree of vertices unchanged but the ranking can change as a result of the estimation method.
An interesting question is 'How an additional match with a given result affects the ranking?' 
The impact of lower and higher values of sparsity (degrees of vertices) can also be analysed. A challenging question could be 'Which player can be cancelled without changing the ranking?'

The inconsistency of pairwise comparison matrices plays an important role both in theory and practice \citep{Keri2011}. Further research includes the analysis of inconsistency of incomplete pairwise comparison matrices of large size. In our case we cannot speak of the inconsistency of a decision maker since the matrix elements originate from tennis matches, we might also say: from life. Intransitive triads ($A$ beats $B$, $B$ beats $C$, and $C$ beats $A$) occur often in sports. We have found $50$ intransitive triads in our example, they are plotted in the Online Appendix, available at \url{http://www.sztaki.mta.hu/~bozoki/tennis/appendix.pdf}.
We hope to return to the problem of analysing intransitive triads in a(n incomplete) pairwise comparison matrix, or, equivalently, in the directed graph associated.

\section*{Acknowledgement}

The authors are grateful to the anonymous reviewers for valuable and constructive recommendations.
Research was supported in part by OTKA grants K 77420 and K 111797.


\end{document}